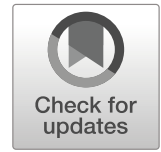

# Vehicle Attribute Recognition by Appearance: Computer Vision Methods for Vehicle Type, Make and Model Classification

Xingyang Ni[1] · Heikki Huttunen[1]

**Abstract**
This paper studies vehicle attribute recognition by appearance. In the literature, image-based target recognition has been extensively investigated in many use cases, such as facial recognition, but less so in the field of vehicle attribute recognition. We survey a number of algorithms that identify vehicle properties ranging from coarse-grained level (vehicle type) to fine-grained level (vehicle make and model). Moreover, we discuss two alternative approaches for these tasks, including straightforward classification and a more flexible metric learning method. Furthermore, we design a simulated real-world scenario for vehicle attribute recognition and present an experimental comparison of the two approaches.

**Keywords** Vehicle attribute recognition · Image classification · Metric learning

## 1 Introduction

Traffic monitoring is an essential tool for collecting statistics to enable better design and planning of transport infrastructure. Often, plain vehicle counting is not enough, and there is a need to capture extended attributes of the vehicles; for example, can separation of heavy traffic from lighter vehicles, or following individual vehicles to find out which routes the drivers usually take? Such data allows more fine-grained analysis and more accurate profiling of users of transport infrastructure, which is necessary for assessing the effects of future changes in transportation.

The traditional approach for collecting traffic data is to organize manual data collection campaigns and enroll human labor for counting the vehicles, or implement roadside questionnaires after stopping the vehicles. Needless to say, such labor-intensive operations are quite insulting to drivers' experience. Alternatively, various technologies could be used to facilitate the data collection procedure.

The inductive ground loops (see, e.g., MAVE®-L product line of AVE GmbH)[1] measure the magnetic profile of vehicles passing by and provide a coarse-grained classification of vehicle type. Moreover, laser scanners (see, e.g., traffic counters from SICK GmbH)[2] can gather similar information from vehicles. Even audio can be applied as a means of identifying the type of a vehicle by extracting the predefined feature set from segments of short audio signal [75].

More recently, camera-based techniques for traffic monitoring have become more widespread. Cameras are ubiquitous, cost-effective, and often have been already utilized in other surveillance use cases. Moreover, they provide a rich source of information with which new generation of recognition methods become feasible; for example, inductive ground loops and laser scanners enable only a very coarse-grained categorization due to the nature of the input data they have. The level of computer vision techniques was the bottleneck of camera-based traffic monitoring systems for a long time. However, the emergence of deep learning has thoroughly changed the situation. In particular, image classification has progressed to an entirely new level within the last ten years and is reaching human-level accuracy in many domains. An essential factor in this transformation is the availability of large-scale datasets. Significant milestones in the history of such datasets include the

✉ Xingyang Ni
xingyang.ni@tuni.fi

Heikki Huttunen
heikki.huttunen@tuni.fi

[1] Tampere University, Tampere, Finland

---

[1]http://www.ave-web.de/
[2]http://www.sick.de/

ImageNet dataset for image classification [9], the Microsoft COCO dataset for object detection [42], and more recent vehicle-specific datasets such as the KITTI dataset for autonomous driving [18], the VERI-Wild dataset for vehicle re-identification [48], and the CCPD dataset for license plate recognition [79].

Deep learning techniques have enabled diverse practical applications of vehicle attribute recognition. In this paper, we will discuss the following three problem settings:

- **Vehicle type recognition** attempts to characterize vehicles to coarse-grained categories by their size or intended usage, e.g., sedan, bus and truck. A common use case is statistical: What is the distribution of vehicle categories at a specific checkpoint?
- **Vehicle make recognition** categorizes vehicles by their manufacturer, e.g., Ford, Toyota and Chevrolet. An example use case is searching for stolen vehicles or license plates: Did the make of this vehicle change since last month?
- **Vehicle model recognition** learns to predict the vehicle model, e.g., Ford Puma, Toyota Corolla and Chevrolet Volt. This task is significantly more detailed than the aforementioned vehicle make recognition, and it can be used as a method to study consumer behavior regarding a specific vehicle model. Additional challenges are introduced due to the dynamic nature of the data: Manufacturers introduce new designs annually, and the prediction model needs to be regularly updated.

Additional topics outside the scope of this paper include vehicle re-identification [45, 47, 48] which is an exciting and prevalent area that is related to metric learning described below. It targets to match vehicles across multiple cameras based on their appearance. Such methods allow vehicle tracking in larger regions, which helps in planning the road network. Alternatively, one could use license plate recognition for the same task. However, re-identification does not explicitly require the license plate to be visible and allows a less restricted camera placement (e.g., longer distance, or non-frontal camera angle).

The benefit of type, make and model recognition is that they do not collect any privacy-sensitive information about individual vehicles. The results are purely statistical (e.g., "15% of traffic on this road are buses"), and may thus avoid potential privacy issues. On the other hand, license plate recognition and re-identification systems can also be implemented in a privacy-preserving manner, where only a hash of the recognized license plate or a feature vector representing the appearance is retrieved, rather than a image of the vehicle itself. Nevertheless, their use may be still be less acceptable by the public than a purely statistical approach.

There are two commonly used strategies for type, make and model recognition. First is the straightforward approach that poses the recognition task as a classification problem. For example, vehicle type recognition typically has only a few distinct classes with abundant samples of each class. It is natural to present it as a $K$-way classification problem [27, 36].

The classification approach may not be feasible for all cases. In re-identification, it is difficult to train a classification model since the large number of identities would make the last fully-connected layer excessively huge. Therefore, this problem is usually approached as a metric learning problem, where a neural network learns a mapping function from images to feature vectors. The identity can be retrieved by comparing the feature vectors against the historical collection of feature vectors, most often applying the nearest neighbor search method. Another significant benefit of the metric learning approach is that it allows addition of new data (e.g., new car models when they are launched) without retraining the entire model. With this in mind, we will describe both approaches, and carry out experimental analyses of their performance.

The rest of this paper is organized as follows. In Section 2, we discuss available datasets and different problem settings of vehicle attribute recognition. Section 3 explains the methodology of our proposed solution to vehicle type and make recognition, and reports the experimental results on the VERI-Wild [48] dataset. Section 4 describes the future research directions. Finally, Section 5 contains the concluding remark.

## 2 Vehicle Attribute Recognition

In most problems in computer vision, methods can be roughly divided into two categories: older *Hand-crafted feature engineering* methods and newer *Deep learning* approaches. We will discuss these next.

Hand-crafted methods rely on human-engineered feature extraction pipelines to transform the image into a set of features that are robust to variations in both vehicle-specific variables (e.g., scale, location and color), as well as environment variables (e.g., pose, illumination and background). The feature extraction stage is followed by a conventional machine learning (*i.e.*, non-deep learning) classifier, such as nearest neighbor search or Support Vector Machine (SVM) [6].

Deep learning methods differ from the hand-crafted methods in that they do not require human-engineered feature extractors, but instead learn the feature extractors purely from data. This is the main reason for clearly superior accuracy compared to traditional approaches. For a more in-depth discussion of deep learning, we refer the reader to [15, 19, 26].

### 2.1 Datasets

Image-based vehicle attribute recognition has been studied extensively, especially since the introduction of deep learning techniques. Table 1 lists commonly used datasets on vehicle attribute recognition, sorted by publication year. In addition to the dataset size, the number of unique types/makes/models is shown when applicable.

Historically, the earliest dataset was collected by Petrovic and Cootes [55], and it concentrated on vehicle model recognition. The dataset contains 1,132 frontal images from 77 distinct vehicle models. Likewise, Clady et al. [5] introduced a set of frontal vehicle images with annotations of models. More recently, the car-types [67] dataset has doubled the number of images among the early datasets, with images captured from different viewpoints.

Later, Peng et al. [54] built up a collection of high-resolution frontal images taken under daylight and nightlight, with annotated labels of vehicle type. Both BMW-10 and car-197 datasets were proposed in [39], with the former containing an ultra-fine-grained dataset of 10 BMW sedans and the latter comprising significantly more images from 197 models. Interestingly, the FG3DCar [44] dataset is annotated with additional 64 landmark locations, which makes it feasible to apply 3D-based methods.

In 2015 and 2016, vehicle attribute recognition has seen a surge of interest as many new datasets [13, 27, 41, 64, 80] have been gathered. The most notable datasets are CompCars [80] and BoxCars [64], which contain a large number of images with fine-grained labels of specific vehicle models. The CompCars dataset consists of images collected from either public websites or surveillance cameras, enabling real-world applications that need to address the challenge of significant appearance variations. On the other hand, the BoxCars dataset focuses on traffic surveillance applications and includes a 3D bounding box for each vehicle and the foreground mask.

As might be expected, the dataset scale is continuously increasing. The updated BoxCars116k [65] dataset extends the previous BoxCars [64] to contain almost twice the number of images. The MIO-TCD [51] dataset is a significant milestone pushing vehicle attribute recognition to the next level. The dataset is many times larger than any previous datasets, and it features a localization subset for the detection task and a classification subset for the recognition task.

Among the most recent datasets, the VERI-Wild [48] dataset was initially collected for vehicle re-identification, and images are captured under unconstrained scenarios. Nevertheless, the provided vehicle type and make annotations make it also suitable for recognition purposes.

It stands out to argue that deep neural networks benefit considerably from large-scale datasets. For future research, the MIO-TCD [51] dataset, the VERI-Wild [48] dataset and the CompCars [80] dataset are well suited to vehicle type recognition, vehicle make recognition and vehicle model recognition, respectively.

### 2.2 Methods for Vehicle Type Recognition

Vehicle type recognition aims at a coarse-grained prediction of vehicle type, with popular categories including sedan, bus and truck. Table 2 summarizes conspicuous hand-crafted and deep learning methods for vehicle type recognition. Note that the accuracies of different methods are not comparable if they are not tested on the same dataset.

**Table 1** Commonly used datasets on vehicle attribute recognition, sorted by publication year.

| Dataset | Year | #Image | #Type | #Make | #Model |
|---|---|---|---|---|---|
| Petrovic and Cootes [55] | 2004 | 1,132 | – | – | 77 |
| Clady et al. [5] | 2008 | 1,121 | – | – | 50 |
| car-types [67] | 2011 | 1,904 | – | – | 14 |
| Peng et al. [54] | 2012 | 4,924 | 5 | – | – |
| BMW-10 [39] | 2013 | 512 | 1 | 1 | 10 |
| car-197 [39] | 2013 | 16,185 | 7 | – | 197 |
| FG3DCar [44] | 2014 | 300 | – | – | 30 |
| Liao et al. [41] | 2015 | 1,482 | – | 8 | – |
| BIT-Vehicle [13] | 2015 | 9,850 | 6 | – | – |
| CompCars [80] | 2015 | 214,345 | 12 | 161 | 1,687 |
| Huttunen et al. [27] | 2016 | 6,555 | 4 | – | – |
| BoxCars [64] | 2016 | 63,750 | – | 27 | 148 |
| BoxCars116k [65] | 2018 | 116,286 | – | 45 | 693 |
| MIO-TCD [51] | 2018 | 648,959 | 11 | – | – |
| VERI-Wild [48] | 2019 | 416,314 | 14 | 149 | – |

**Table 2** Selected hand-crafted and deep learning methods for vehicle type recognition. Algorithms tested on the same dataset are grouped.

| Method | Year | Dataset | #Image | #Type | Accuracy | Notes |
|---|---|---|---|---|---|---|
| Petrovic and Cootes [54, 55] | 2004 | Peng et al. [54] | 4,924 | 5 | 84.3% | Hand-crafted |
| Psyllos et al. [54, 57] | 2011 | | | | 78.3% | Hand-crafted |
| Peng et al. [54] | 2012 | | | | 90.0% | Hand-crafted |
| Dong and Jia [12] | 2013 | | | | 91.3% | Hand-crafted |
| Peng et al. [53] | 2013 | | | | 93.7% | Hand-crafted |
| Dong et al. [13] | 2015 | | | | 96.1% | Deep learning |
| Kafai and Bhanu [33] | 2012 | Kafai and Bhanu [33] | 845 | 4 | 96.6% | Hand-crafted |
| Petrovic and Cootes [2, 55] | 2004 | BIT-Vehicle [13] | 9,850 | 6 | 78.6% | Hand-crafted |
| Psyllos et al. [2, 57] | 2011 | | | | 70.8% | Hand-crafted |
| Peng et al. [2, 53] | 2013 | | | | 85.0% | Hand-crafted |
| Dong et al. [13] | 2015 | | | | 88.1% | Deep learning |
| Sun et al. [68] | 2017 | | | | 90.1% | Hand-crafted |
| Yang et al. [80] | 2015 | Subset of CompCars [80] | 52,083 | 12 | 63.1% | Deep learning |
| Huttunen et al. [27] | 2016 | Huttunen et al. [27] | 6,555 | 4 | 97.8% | Deep learning |
| He et al. [22, 58] | 2015 | MIO-TCD [51] | 648,959 | 11 | 96.5% | Deep learning |
| Kim and Lim [37] | 2017 | | | | 97.8% | Deep learning |
| Lee and Chung [71] | 2017 | | | | 97.9% | Deep learning |
| Jung et al. [31] | 2017 | | | | 97.9% | Deep learning |
| Theagarajan et al. [72] | 2017 | | | | 97.8% | Deep learning |
| Rachmadi et al. [58] | 2018 | | | | 97.9% | Deep learning |

### 2.2.1 Hand-Crafted Methods

Psyllos et al. [57] defined the Region of Interest based on the size and location of the license plate after recognizing the plate with the Sliding Concentric Window segmentation method [8]. A Probabilistic Neural Network [66] is trained on a set of Scale Invariant Feature Transform (SIFT) [49] feature descriptors, and it accelerates inference speed considerably compared with the conventional nearest neighbor classifier.

Peng et al. have done a series of works in vehicle type recognition. In [54], a coarse-to-fine method is proposed to enable fast and accurate license plate localization. A coarse-grained detection is obtained by inspecting the intensity histograms horizontally, while the line segments feature generates finer localization. Eigenvectors are extracted from vehicle front images as the feature representation. K-mean clustering is applied to each vehicle type, and the category of a test sample is in line with the nearest cluster center. Later on, the aforementioned method gets improved in three aspects [53]. First, the color of the license plate is an informative clue to vehicle type, and it is incorporated into the classification pipeline. Furthermore, the coordinates of the vehicle are extracted accurately with a straightforward background-subtraction method. Lastly, the algorithm estimates the vehicle type based on the top ten most similar training samples rather than only the best match.

In [12], two sets of feature embedding are utilized. On the one hand, the SIFT [49] feature analyzes local patterns in images, and it falls into the category of appearance-based features. On the other hand, the relative coordinate between each SIFT keypoint and the mean keypoint of a local region correspond to the structural feature. A Multiple Kernel Learning method is proposed to merge the two feature sets mentioned above and generate a more robust prediction.

Comparably, Sun et al. [68] extracted two sets of feature embedding individually. The global feature set is produced by an improved Canny edge detection algorithm, while the local feature set is extracted from by applying Gabor wavelet kernels on non-overlapping patches of the whole vehicle image. In addition, a two-stage classification strategy is proposed: the first stage model predicts whether the sample is a small vehicle or a large vehicle while the second stage model recognizes the specific vehicle type.

Kafai and Bhanu [33] shifted focus on video-based vehicle type recognition from direct rear-side view for the reason that a vehicle does not necessarily have a front license plate. After detecting a moving vehicle with a Gaussian mixture model, the coordinates of the license plate are extracted by using either a matched filtering approach [1] which exploits the colored texture in the plate, or a blob detection and filtering method which picks out the best match from candidate blobs. In addition to the license plate, the regions of tail light are located by examining the redness of pixels. A low-level feature set is computed from each frame, e.g., height, width, and angle of tail lights. Subsequently, a Hybrid Dynamic Bayesian Network is implemented by adding an explicit temporal dimension to a standard Bayesian Network, and it generates a probability distribution with respect to the feature vectors.

### 2.2.2 Deep Learning Methods

Dong et al. [13] adopts a semi-supervised convolutional neural network to classify vehicle type. The neural network consists of convolutional layers, absolute value rectification layers, local contrast normalization layers, average pooling layers, subsampling layers, and a fully-connected layer with softmax activation. A sparse Laplacian filter learning method is proposed to optimize the parameters of convolutional layers with a large number of unlabeled samples. On the contrary, the parameters in the fully-connected layer are learned on labeled samples.

Huttunen et al. [27] compared the performances of a deep learning neural network with a hand-crafted method which employs SVM [6] on SIFT [49] feature. Instead of using either manual or grid search of the optimal setting of neural network's hyperparameters such as input image size, kernel size of convolutional layers and learning rate, the random search strategy [3] reduces computational burden significantly while reaching comparable or even superior performance. The resulting topology outperforms the SVM classifier in terms of accuracy.

Since the introduction of the large-scale MIO-TCD [51] dataset, deep learning has become the predominant approach for vehicle type recognition. Kim and Lim [37] choose a convolutional neural network of moderate size, and the samples are augmented with flipping and rotations. Multiple models are trained while each of them accesses a half portion of the training set, which is randomly selected. Consequently, the classification system produces several predictions for a single test sample. The results are aggregated by a voting process that imposes different weights to each class label so that the problem of imbalanced data is compensated.

Lee and Chung [71] propose an ensemble of 12 local expert networks and 6 global networks. The local expert networks take the GoogLeNet [69] structure, and each network is trained on a subset of training samples. The dataset is split in view of the resolution and aspect ratio of samples. Conversely, the global networks are trained with all training samples and three topologies are used, namely, AlexNet [40], GoogLeNet [69] and ResNet [23]. In the inference procedure, a rule-based gating function [29] selects the prediction from a specific local expert network considering the resolution and aspect ratio of the test sample. The final prediction is generated by merging the predictions of single local expert network and multiple global networks.

Jung et al. [31] train ResNet [23] models on samples augumented by photometric distortions [24] and color modifications [40]. Multiple ResNet-based backbones are optimized simultaneously while their outputs are element-wise added up. A joint fine-tuning method [32] is employed to fine-tune all parameters rather than only the last dense layer. Besides, a mechanism named DropCNN randomly drops the predictions from the aforementioned backbones during training.

In [72], two neural networks are trained independently with the weighted cross-entropy loss function. Both models are based on ResNet [23], and they differ in the number of layers. The logical reasoning is appended to the fully-connected layer to confront the issue of dual-class misclassification. The predictions of different models are combined using weights, which refer to the average values of precision and recall.

Last but not least, Rachmadi et al. [58] introduces a Pseudo Long Short-Term Memory (P-LSTM) classifier for identifying a single image. Unlike the ordinary use cases which involve time-series data, multiple parallel networks extract features from different crops of the input image, and those spatial pyramid features are feed to the P-LSTM classifier in sequence. A fully-connected layer is appended at the end to compute the probabilities of each class label.

### 2.3 Methods for Vehicle Make and Model Recognition

Vehicle make recognition targets to predict the manufacturer or brand of the vehicle (e.g., Ford, Toyota or Chevrolet). By contrast, vehicle model recognition aims at a more fine-grained prediction of the particular model (e.g., Ford Puma, Toyota Corolla or Chevrolet Volt). The characteristics of type recognition differ from make and model recognition in two aspects. The number of categories in type recognition is significantly smaller than that in make and model recognition. Besides, type recognition tends to have static categories, whereas makes and models change at times. Consequently, type recognition is commonly considered as a more manageable task where a straightforward classification setup and even hand-crafted classifiers can be competitive.

Tables 3 and 4 compile prominent hand-crafted and deep learning methods for vehicle make recognition and vehicle model recognition, respectively. Note that the accuracies of different methods are not comparable if they are not tested on the same dataset.

**Table 3** Selected hand-crafted and deep learning methods for vehicle make recognition. Algorithms tested on the same dataset are grouped.

| Method | Year | Dataset | #Image | #Make | Accuracy | Notes |
| --- | --- | --- | --- | --- | --- | --- |
| Liao et al. [41] | 2015 | Liao et al. [41] | 1,482 | 8 | 81.3% | Hand-crafted |
| Yang et al. [80] | 2015 | | | | 82.9% | Deep learning |
| Hu et al. [25] | 2017 | Subset of CompCars [80] | 30,955 | 75 | 99.3% | Deep learning |
| Xiang et al. [76] | 2019 | | | | 99.6% | Deep learning |

### 2.3.1 Hand-Crafted Methods

In the early works, methods are typically constrained to vehicle images captured from the frontal view, and the nearest neighbor search is employed to find the most similar sample. In [55], the license plate is detected by searching for all possible right angle corners and selecting the best candidate while considering the scale and aspect constraints of its rectangle structure. A set of structure mapping methods are investigated to extract the feature vector from the Region of Interest, e.g., raw pixel values, Harris corner detector [21], and square mapped gradients. Clady et al. [5] propose to construct a model from several frontal vehicle images based on oriented-contour points. Given a prototype image, the oriented-contour points matrix is computed by applying a histogram-based threshold process on the gradient orientations. A discriminant function measures the similarity scores between test samples and labels in trained models.

Later methods take advantage of the Deformable Part Model (DPM) [14] algorithm, which is originally proposed to solve generic object detection tasks. Stark et al. [67] suggest that the detected parts indicate the geometry of objects and help in matching the class labels. The vanilla DPM is reformulated as a latent linear multi-class SVM [6], and the consequential structDPM classifier is directly optimized against a multi-class loss function. Liao et al. [41] construct the hypothesis that vehicle parts differ in the discriminative capacity of estimating vehicle attributes. After a DPM-based detector localizes vehicle parts, multiple predictions are generated based on the

**Table 4** Selected hand-crafted and deep learning methods for vehicle model recognition. Algorithms tested on the same dataset are grouped.

| Method | Year | Dataset | #Image | #Model | Accuracy | Notes |
| --- | --- | --- | --- | --- | --- | --- |
| Petrovic and Cootes [55] | 2004 | Petrovic and Cootes [55] | 1,132 | 77 | 93% | Hand-crafted |
| Clady et al. [5] | 2008 | Clady et al. [5] | 1,121 | 50 | 93.1% | Hand-crafted |
| Stark et al. [67] | 2011 | car-types [67] | 1,904 | 14 | 93.5% | Hand-crafted |
| Krause et al. [39] | 2013 | | | | 94.5% | Hand-crafted |
| Krause et al. [39] | 2013 | BMW-10 [39] | 512 | 10 | 76.0% | Hand-crafted |
| Krause et al. [39] | 2013 | | | | 67.6% | Hand-crafted |
| Krause et al. [38] | 2015 | car-197 [39] | 16,185 | 197 | 92.8% | Deep learning |
| Hu et al. [25] | 2017 | | | | 93.1% | Deep learning |
| Xiang et al. [76] | 2019 | | | | 94.3% | Deep learning |
| Lin et al. [44] | 2014 | FG3DCar [44] | 300 | 30 | 90.0% | Hand-crafted |
| Yang et al. [80] | 2015 | | | | 76.7% | Deep learning |
| Hu et al. [25] | 2017 | Subset of CompCars [80] | 52,083 | 431 | 97.6% | Deep learning |
| Xiang et al. [76] | 2019 | | | | 98.5% | Deep learning |
| Jaderberg et al. [30, 81] | 2015 | | | | 64.3% | Deep learning |
| Sochor et al. [64] | 2016 | Subset of BoxCars [64] | 59,742 | 77 | 75.4% | Deep learning |
| Fu et al. [16, 81] | 2017 | | | | 72.2% | Deep learning |
| Zeng et al. [81] | 2019 | | | | 81.2% | Deep learning |
| Lin et al. [43, 65] | 2015 | | | | 69.6% | Deep learning |
| Simon and Rodner [63, 65] | 2015 | Subset of BoxCars116k [65] | 90,840 | 107 | 75.9% | Deep learning |
| Gao et al. [17, 65] | 2016 | | | | 70.6% | Deep learning |
| Sochor et al. [65] | 2018 | | | | 84.1% | Deep learning |

Histograms of Oriented Gradient (HOG) [7] features of each part. Those predictions are accumulated with a weighting scheme allowing that larger weights are assigned to more influential parts.

Compared with the works explained above, some researchers divert the attention to 3D space. Krause et al. [39] extend two methods to obtain superior object representations in 3D. Established on the basis of 2D Spatial Pyramid [73], each rectified patch is associated with corresponding 3D coordinates. Likewise, the pooling regions in 2D BubbleBank [11] is switched from 2D to 3D. The resulting 3D representations from these two methods are combined with linear SVM [6] classifiers, which are trained in the manner of one-versus-all.

In [44], Lin et al. optimize 3D model alignment and fine-grained classification jointly. To begin with, the DPM method gives a rough estimation of part locations, while a pre-trained regression model further detects the representative landmarks. Each 3D model consists of a collection of 3D points, and the 3D object geometry is obtained by fitting the model to landmark locations in 2D. With hand-crafted features retrieved from each landmark, a multi-class linear SVM [6] classifier predicts the label. Most importantly, the predicted label is fed back to the 3D model to get better alignment results.

### 2.3.2 Deep Learning Methods

The CompCars [80] dataset contains a considerable number of vehicle images taken from all viewpoints with rich annotations. Yang et al. [80] utilize an Overfeat model [61] which is initialized with pretrained weights on the ImageNet [9] dataset, and fine-tune it for vehicle attribute recognition. The Overfeat model differs from the established AlexNet model [40] in three aspects: (i) it does not contain a contrast normalization scheme; (ii) adjacent pooling regions do not overlap and (iii) smaller stride value is used to get larger feature maps.

Hu et al. [25] surpass the previous work considerably on the strength of a novel spatially weighted pooling scheme. The pooling layer learns spatially weighted masks which assess the discriminative capacity of spatial units, and applies pooling operation to the extracted feature maps of the convolutional layer correspondingly.

Xiang et al. [76] propose a four-stage pipeline that takes the interaction between parts into account. Part detection is implemented using a backbone model truncated at an intermediate layer, while part assembling involves pointwise convolutional layers which gather associated parts into the same feature map. Afterward, topology constraint comprises depthwise convolutional layers and estimates the probability of the topology relationship between related parts. The ending classification uses a fully-connected layer to make predictions.

Sochor et al. are noted for collecting the BoxCars [64] and BoxCars116k [65] datasets. In [64], the recognition performance is boosted by inserting additional supplementary information to the neural network, more specifically, 3D vehicle bounding box, rasterized low-resolution shape, and 3D vehicle orientation. With 3D bounding boxes automatically obtained from surveillance cameras and the rasterized low-resolution shape information, a normalization procedure aligns the original images. Besides, the 3D orientation offers an insight into the viewpoint, which is beneficial.

Later on, the previous method is extended in [65] by proposing a method to estimate 3D bounding boxes in case of such information is unavailable. The directions to the vanishing points are obtained from three classification branches, which generate probabilities for each vanishing point belonging to a specific angle. In addition, two extra data augmentation strategies are integrated. On the one hand, the color of the image is randomly alternated. On the other hand, random crops in the images are filled with random noise.

Zeng et al. [81] devise a framework that learns a joint representation of the 2D global texture and 3D bounding box. The 2D global feature originates from a pre-trained detector that localizes the Region of Interest. The 3D perspective network regresses the 3D bounding box and extracts feature embedding. At last, the feature fusion network merges the set of two features and generates the predictions.

Unlike the aforementioned methods which are explicitly devised for vehicle make and model recognition, some generic image classification algorithms have also been validated, especially from the domain of fine-grained image classification. Jaderberg et al. [30] introduce a differentiable spatial transformer module, which makes the trained models more spatially invariant to the input data. A spatial transformer spatially transforms feature maps, and the manipulation is conditioned on the feature maps itself. The localisation network regresses the transformation parameters. The grid generator produces a set of points where the input feature maps should be sampled. Finally, the sampler samples the input feature maps at the grid points.

In [63], Simon and Rodner present a method that learns part models without the need of acquiring annotations of parts or bounding boxes. The channels in feature maps are treated as a part detector, and a part constellation model is obtained by selecting part detectors that fire at similar relative locations. As a side product, the filtered part proposals can be applied in the data augmentation pipeline.

Lin et al. [43] utilize a bilinear model consisting of two feature extractors. The feature vectors from those extractors are multiplied using the outer product and pooled to obtain the bilinear vector. The resulting topology is significantly faster than methods that are dependent on detectors, and it is capable of capturing pairwise correlations between the feature channels.

From a different perspective, Gao et al. [17] targeted on reducing the size of feature representation in bilinear models, without compromising the discriminative capacity. Two compact bilinear pooling approaches are investigated to approximate the inner product of two descriptors, namely, Random Maclaurin [34] and Tensor Sketch [56]. The compact pooling methods are differentiable so that the pipeline can be optimized end-to-end.

Last but not least, Fu et al. [16] propose to recursively learn discriminative region attention and region-based feature representation at multiple scales. The classification sub-network provides sufficient discriminative capacity at each scale. The attention proposal sub-network starts from the full image, and iteratively generates region attention from coarse to fine. Meanwhile, the finer scale network takes a magnified region from previous scales in a recurrent manner.

# 3 Experimental Implementation of a Vehicle Attribute Recognizer

As discussed in Section 2.3, vehicle make and model recognition have been extensively studied, but there is still room for further investigation, especially due to the recently introduced large-scale datasets. Namely, especially people recognition has recently taken significant advances both in facial and full-body recognition, and some recent pipelines have not been experimented on vehicles yet. Therefore, we will next revisit a state-of-the-art method for people re-identification [50] and experiment on its out-of-the-box accuracy in the domain of vehicle attribute recognition. Moreover, the experiment also aims to present the details of implementing a vehicle identification model in a concrete manner.

## 3.1 Methodology

### 3.1.1 Model Architecture

The backbone model ResNet50 [23] is initialized with pre-trained weights on ImageNet [9]. The global average pooling layer shrinks the spatial dimensions of feature maps and generates a feature vector for each sample. On the condition that the categorical cross-entropy loss function is applied, a batch normalization [28] layer and a fully-connected layer are appended to the end so that the model predicts the probabilities of classes.

### 3.1.2 Loss Function

Three loss functions are examined, namely, categorical cross-entropy loss, triplet loss [60] and lifted structured loss [52]. The categorical cross-entropy loss function is widely used in conventional classification problems. The last two falls in the scope of metric learning, which optimizes feature embedding directly in such a way samples within the same class get closer while samples from different classes get further.

### 3.1.3 Training Procedure

Four strategies are implemented to boost performance. In the early stage of the training, the learning rate starts from a low value and increases gradually as the training proceeds. Such a warmup strategy cracks the distorted gradient issue [20, 46]. Moreover, random erasing data augmentation [83] is utilized to mask out random crops from original images, and it helps the model generalize better. Finally, label-smoothing regularization [70] encourages the model to make less confident predictions, and $\ell_2$ regularization induces the model to choose smaller parameters.

### 3.1.4 Inference Procedure

Given a test sample, the output of the global average pooling layer is retrieved as the feature embedding. The cosine distance function is chosen to measure the distance between two feature vectors. While the entire test set corresponds to the query set, 100 samples are randomly selected from each class label in the training set and those samples constitute the gallery set. Each query sample takes the label of the closest match from the gallery set, and accuracy measures the percentage of cases in which the predicted label is consistent with the ground truth label.

The reason why we adopt the paradigm as mentioned above is that it is also applicable in cases with dynamic content. In the matter of a real-world vehicle model recognition system, automobile manufacturers release new car models regularly. The conventional classification approach, which computes the probabilities of classes that are available in the training set, is doomed to fail on unseen car models. In contrast, building up a gallery set and comparing the test sample with gallery samples might still yield a meaningful prediction. Whenever new car models get released, a certain amount of exemplary images could be added to the gallery set without the need of retraining the neural network itself.

## 3.2 Experimental Results

Table 5 reports the accuracy of categorical cross-entropy loss, triplet loss [60] and lifted structured loss [52] on

**Table 5** Experimental results on the VERI-Wild [48] dataset which contains approximately 0.4 million images.

| Method | #Type | Accuracy on Type | #Make | Accuracy on Make |
|---|---|---|---|---|
| cross-entropy loss |  | 96.8% |  | 95.6% |
| triplet loss | 14 | 97.4% | 149 | 95.3% |
| lifted structured loss |  | 97.7% |  | 96.2% |

Test performances of categorical cross-entropy loss, triplet loss [60] and lifted structured loss [52] are evaluated on vehicle type and make recognition.

two tasks, *i.e.*, vehicle type and make recognition. It is explicit that lifted structured loss achieves superior performance than the other two loss functions. Therefore, incorporating the latest advances in metric learning, e.g., ArcFace loss [10], might push the boundaries of vehicle attribute recognition even further.

# 4 Future Research Directions

In this section, we point out four future research directions, namely, few-shot learning, multi-task learning, attention, and edge computing. Those topics are under-developed in the domain of vehicle attribute recognition.

## 4.1 Few-Shot Learning

New car models hit the consumer market from time to time. This fact poses a challenging problem, especially for any real-world vehicle model recognition system. Without adopting an appropriate solution, the classifier could not identify new car models. In the area of few-shot learning, a model has to recognize new classes with only a few samples from those unseen classes [74]. Setting up a vehicle attribute recognition pipeline in the manner of few-shot learning would make the algorithm more applicable in practice.

## 4.2 Multi-task Learning

Multi-task learning refers to the learning procedure in which multiples tasks are optimized simultaneously [59]. In ideal cases, the model achieves better generalization by reason of sharing feature representations across relevant tasks [35]. Some vehicle attribute recognition datasets provide several related attributes, e.g., CompCars [80], BoxCars [64] and VERI-Wild [48]. The availability of such annotations makes multi-task learning a feasible topic on vehicle attribute recognition.

## 4.3 Attention

Following the pattern of the human visual system, the attention mechanism allows the neural network to pay attention to certain salient parts of the input dynamically [78]. In terms of fine-grained image classification, the attention mechanism assists the model in spotting subtle visual differences between distinct categories and obtaining better performance [16, 77, 82]. It is anticipated that automobile manufacturers make incremental modifications to a specific model and introduce a variant model [80]. Therefore, the attention mechanism is beneficial, especially in hard cases, *i.e.*, different vehicle models from the same product series.

## 4.4 Edge Computing

While the cameras set up on the highway capture video streams, one could deploy the vehicle attribute recognition algorithms directly on edge devices. The concept of edge computing can be of great benefit in shorter response time, lower bandwidth cost, and safer data security [62]. Since edge devices are typically resource-constrained, lightweight models are preferable while balancing speed and accuracy. Possible solutions include parameter pruning and sharing, low-rank factorization, transferred/compact convolutional filters, and knowledge distillation [4].

# 5 Conclusion

In this paper, we have surveyed the state-of-the-art vehicle attribute recognition algorithms both for coarse-grained (vehicle type) and fine-grained (vehicle make and model) attributes. Since the advent of deep learning techniques, the recognition accuracy has taken significant leaps and is already likely to exceed human-level performance. The unforeseen progress enables new kinds of applications, that earlier low accuracy techniques did not facilitate.

In addition to the hardware resources which support efficient parallelization, another critical enabler for high accuracy is large-scale datasets. We listed an extensive collection of datasets for the three tasks and discussed their strengths and weaknesses. Those datasets contain considerable variations: some of them only have a fixed pose, fixed illumination, or a minimal set of categories; while others are taken "in the wild", and the challenges brought by the unconstrained environment variables are demanding. In real-world applications, the situation is usually something between these extremes, and we believe that this survey helps in searching for suitable datasets and mixing them for maximum performance.

On the algorithms side, we discussed two typical training setups: classification setup (where each type/make/model is one class) and metric learning setup (where the model seeks to learn a distance function). The former approach is simple to comprehend and easy to implement. However, the challenge is the static nature of the classification model: if a new class appears, the model needs to be adapted. On the other hand, the metric learning approach may have a steeper learning curve, but it can offer increased flexibility and superior performance.

The classification approach is most suitable for cases where the classes are unvarying; for example, the unique vehicle types are unlikely to change any time soon. By contrast, more dynamic cases include vehicle make and model recognition, where new categories appear annually. Metric learning is preferable in such cases, while a test sample is compared to a gallery of examples using the nearest neighbor search method.

Finally, we compared the two approaches quantitatively for vehicle type and make recognition, and discovered that the metric learning approach with properly selected loss function outperforms the classification approach by a clear margin. This novel comparison based on a competitive pipeline is enlightening for practitioners, and it highlights the promising potential of metric learning, which has seen significant advances during recent years.

**Acknowledgements** This work was financially supported by Business Finland project 408/31/2018 MIDAS.

# References

1. Abolghasemi, V., & Ahmadyfard, A. (2009). An edge-based color-aided method for license plate detection. *Image and Vision Computing*, *27*(8), 1134–1142.
2. Bai, S., Liu, Z., Yao, C. (2018). Classify vehicles in traffic scene images with deformable part-based models. *Machine Vision and Applications*, *29*(3), 393–403.
3. Bergstra, J., & Bengio, Y. (2012). Random search for hyper-parameter optimization. *The Journal of Machine Learning Research*, *13*(1), 281–305.
4. Cheng, Y., Wang, D., Zhou, P., Zhang, T. (2017). A survey of model compression and acceleration for deep neural networks. arXiv:1710.09282.
5. Clady, X., Negri, P., Milgram, M., Poulenard, R. (2008). Multi-class vehicle type recognition system. In *IAPR workshop on artificial neural networks in pattern recognition* (pp. 228–239): Springer.
6. Cortes, C., & Vapnik, V. (1995). Support-vector networks. *Machine Learning*, *20*(3), 273–297.
7. Dalal, N., & Triggs, B. (2005). Histograms of oriented gradients for human detection. In *IEEE computer society conference on computer vision and pattern recognition, 2005. CVPR 2005*, (Vol. 1 pp. 886–893): IEEE.
8. Deb, K., Chae, H.U., Jo, K.H. (2009). Vehicle license plate detection method based on sliding concentric windows and histogram. *Journal of Computers*, *4*(8), 771–777.
9. Deng, J., Dong, W., Socher, R., Li, L.J., Li, K., Fei-Fei, L. (2009). Imagenet: a large-scale hierarchical image database. In *IEEE conference on computer vision and pattern recognition* (pp. 248–255): IEEE.
10. Deng, J., Guo, J., Xue, N., Zafeiriou, S. (2019). Arcface: additive angular margin loss for deep face recognition. In *Proceedings of the IEEE conference on computer vision and pattern recognition* (pp. 4690–4699).
11. Deng, J., Krause, J., Fei-Fei, L. (2013). Fine-grained crowdsourcing for fine-grained recognition. In *Proceedings of the IEEE conference on computer vision and pattern recognition* (pp. 580–587).
12. Dong, Z., & Jia, Y. (2013). Vehicle type classification using distributions of structural and appearance-based features. In *2013 IEEE international conference on image processing* (pp. 4321–4324): IEEE.
13. Dong, Z., Wu, Y., Pei, M., Jia, Y. (2015). Vehicle type classification using a semi-supervised convolutional neural network. *IEEE Transactions on Intelligent Transportation Systems*, *16*(4), 2247–2256.
14. Felzenszwalb, P.F., Girshick, R.B., McAllester, D., Ramanan, D. (2009). Object detection with discriminatively trained part-based models. *IEEE Transactions on Pattern Analysis and Machine Intelligence*, *32*(9), 1627–1645.
15. Chollet, F. (2017). *Deep Learning with Python*, 1st edn. USA: Manning Publications Co.
16. Fu, J., Zheng, H., Mei, T. (2017). Look closer to see better: recurrent attention convolutional neural network for fine-grained image recognition. In *Proceedings of the IEEE conference on computer vision and pattern recognition* (pp. 4438–4446).
17. Gao, Y., Beijbom, O., Zhang, N., Darrell, T. (2016). Compact bilinear pooling. In *Proceedings of the IEEE conference on computer vision and pattern recognition* (pp. 317–326).
18. Geiger, A., Lenz, P., Stiller, C., Urtasun, R. (2013). Vision meets robotics: the kitti dataset. *The International Journal of Robotics Research*, *32*(11), 1231–1237.
19. Goodfellow, I., Bengio, Y., Courville, A. (2016). *Deep learning*. Cambridge: MIT Press.
20. Goyal, P., Dollár, P., Girshick, R., Noordhuis, P., Wesolowski, L., Kyrola, A., Tulloch, A., Jia, Y., He, K. (2017). Accurate, large minibatch SGD: training ImageNet in 1 hour. arXiv:1706.02677.
21. Harris, C.G., & Stephens, M. (1988). A combined corner and edge detector. In *Alvey vision conference*, (Vol. 15 pp. 10–5244): Citeseer.
22. He, K., Zhang, X., Ren, S., Sun, J. (2015). Spatial pyramid pooling in deep convolutional networks for visual recognition. *IEEE Transactions on Pattern Analysis and Machine Intelligence*, *37*(9), 1904–1916.

J Sign Process Syst

23. He, K., Zhang, X., Ren, S., Sun, J. (2016). Deep residual learning for image recognition. In *Proceedings of the IEEE conference on computer vision and pattern recognition* (pp. 770–778).
24. Howard, A.G. (2013). Some improvements on deep convolutional neural network based image classification. arXiv:1312.5402.
25. Hu, Q., Wang, H., Li, T., Shen, C. (2017). Deep CNNs with spatially weighted pooling for fine-grained car recognition. *IEEE Transactions on Intelligent Transportation Systems*, *18*(11), 3147–3156.
26. Huttunen, H. (2019). Deep neural networks: a signal processing perspective. In *Handbook of signal processing systems* (pp. 133–163): Springer.
27. Huttunen, H., Yancheshmeh, F.S., Chen, K. (2016). Car type recognition with deep neural networks. In *2016 IEEE intelligent vehicles symposium (IV)* (pp. 1115–1120). https://doi.org/10.1109/IVS.2016.7535529.
28. Ioffe, S., & Szegedy, C. (2015). Batch normalization: accelerating deep network training by reducing internal covariate shift. arXiv:1502.03167.
29. Jacobs, R.A., Jordan, M.I., Nowlan, S.J., Hinton, G.E. (1991). Adaptive mixtures of local experts. *Neural Computation*, *3*(1), 79–87.
30. Jaderberg, M., Simonyan, K., Zisserman, A., Kavukcuoglu, K. (2015). Spatial transformer networks. In *Advances in neural information processing systems* (pp. 2017–2025).
31. Jung, H., Choi, M.K., Jung, J., Lee, J.H., Kwon, S., Young Jung, W. (2017). Resnet-based vehicle classification and localization in traffic surveillance systems. In *Proceedings of the IEEE conference on computer vision and pattern recognition workshops* (pp. 61–67).
32. Jung, H., Lee, S., Yim, J., Park, S., Kim, J. (2015). Joint fine-tuning in deep neural networks for facial expression recognition. In *Proceedings of the IEEE international conference on computer vision* (pp. 2983–2991).
33. Kafai, M., & Bhanu, B. (2012). Dynamic bayesian networks for vehicle classification in video. *IEEE Transactions on Industrial Informatics*, *8*(1), 100–109.
34. Kar, P., & Karnick, H. (2012). Random feature maps for dot product kernels. In *Artificial intelligence and statistics* (pp. 583–591).
35. Kendall, A., Gal, Y., Cipolla, R. (2018). Multi-task learning using uncertainty to weigh losses for scene geometry and semantics. In *Proceedings of the IEEE conference on computer vision and pattern recognition* (pp. 7482–7491).
36. Khata, M., Shvai, N., Hasnat, A., Llanza, A., Sanogo, A., Meicler, A., Nakib, A. (2019). Novel context-aware classification for highly accurate automatic toll collection. In *2019 IEEE Intelligent vehicles symposium (IV)* (pp. 1105–1110). https://doi.org/10.1109/IVS.2019.8813866.
37. Kim, P.K., & Lim, K.T. (2017). Vehicle type classification using bagging and convolutional neural network on multi view surveillance image. In *Proceedings of the IEEE conference on computer vision and pattern recognition workshops* (pp. 41–46).
38. Krause, J., Jin, H., Yang, J., Fei-Fei, L. (2015). Fine-grained recognition without part annotations. In *Proceedings of the IEEE conference on computer vision and pattern recognition* (pp. 5546–5555).
39. Krause, J., Stark, M., Deng, J., Fei-fei, L. (2013). 3D object representations for fine-grained categorization. In *Proceedings of the IEEE international conference on computer vision workshops* (pp. 554–561).
40. Krizhevsky, A., Sutskever, I., Hinton, G.E. (2012). Imagenet classification with deep convolutional neural networks. In *Advances in neural information processing systems* (pp. 1097–1105).
41. Liao, L., Hu, R., Xiao, J., Wang, Q., Xiao, J., Chen, J. (2015). Exploiting effects of parts in fine-grained categorization of vehicles. In *IEEE international conference on image processing* (pp. 745–749): IEEE.
42. Lin, T.Y., Maire, M., Belongie, S., Hays, J., Perona, P., Ramanan, D., Dollár, P., Zitnick, C.L. (2014). Microsoft coco: common objects in context. In *European conference on computer vision* (pp. 740–755): Springer.
43. Lin, T.Y., RoyChowdhury, A., Maji, S. (2015). Bilinear cnn models for fine-grained visual recognition. In *Proceedings of the IEEE international conference on computer vision* (pp. 1449–1457).
44. Lin, Y.L., Morariu, V.I., Hsu, W., Davis, L.S. (2014). Jointly optimizing 3D modelv fitting and fine-grained classification. In *European conference on computer vision* (pp. 466–480): Springer.
45. Liu, H., Tian, Y., Yang, Y., Pang, L., Huang, T. (2016). Deep relative distance learning: tell the difference between similar vehicles. In *Proceedings of the IEEE conference on computer vision and pattern recognition* (pp. 2167–2175).
46. Liu, L., Jiang, H., He, P., Chen, W., Liu, X., Gao, J., Han, J. (2019). On the variance of the adaptive learning rate and beyond. arXiv:1908.03265.
47. Liu, X., Liu, W., Ma, H., Fu, H. (2016). Large-scale vehicle re-identification in urban surveillance videos. In *IEEE international conference on multimedia and expo* (pp. 1–6): IEEE.
48. Lou, Y., Bai, Y., Liu, J., Wang, S., Duan, L. (2019). VERI-wild: a large dataset and a new method for vehicle re-identification in the wild. In *Proceedings of the IEEE conference on computer vision and pattern recognition* (pp. 3235–3243).
49. Lowe, D.G. (1999). Object recognition from local scale-invariant features. In *The proceedings of the seventh IEEE international conference on computer vision, 1999*, (Vol. 2 pp. 1150–1157): IEEE.
50. Luo, H., Gu, Y., Liao, X., Lai, S., Jiang, W. (2019). Bag of tricks and a strong baseline for deep person re-identification. In *Proceedings of the IEEE conference on computer vision and pattern recognition workshops* (p. 0).
51. Luo, Z., Branchaud-Charron, F., Lemaire, C., Konrad, J., Li, S., Mishra, A., Achkar, A., Eichel, J., Jodoin, P.M. (2018). MIO-TCD: a new benchmark dataset for vehicle classification and localization. *IEEE Transactions on Image Processing*, *27*(10), 5129–5141.
52. Oh Song, H., Xiang, Y., Jegelka, S., Savarese, S. (2016). Deep metric learning via lifted structured feature embedding. In *Proceedings of the IEEE conference on computer vision and pattern recognition* (pp. 4004–4012).
53. Peng, Y., Jin, J.S., Luo, S., Xu, M., Au, S., Zhang, Z., Cui, Y. (2013). Vehicle type classification using data mining techniques. In *The era of interactive media* (pp. 325–335): Springer.
54. Peng, Y., Jin, J.S., Luo, S., Xu, M., Cui, Y. (2012). Vehicle type classification using PCA with self-clustering. In *IEEE international conference on multimedia and expo workshops* (pp. 384–389): IEEE.
55. Petrovic, V.S., & Cootes, T.F. (2004). Analysis of features for rigid structure vehicle type recognition. In *BMVC*, (Vol. 2 pp. 587–596).
56. Pham, N., & Pagh, R. (2013). Fast and scalable polynomial kernels via explicit feature maps. In *Proceedings of the 19th ACM SIGKDD international conference on knowledge discovery and data mining* (pp. 239–247): ACM.
57. Psyllos, A., Anagnostopoulos, C.N., Kayafas, E. (2011). Vehicle model recognition from frontal view image measurements. *Computer Standards & Interfaces*, *33*(2), 142–151.
58. Rachmadi, R.F., Uchimura, K., Koutaki, G., Ogata, K. (2018). Single image vehicle classification using pseudo long short-term memory classifier. *Journal of Visual Communication and Image Representation*, *56*, 265–274.

59. Ruder, S. (2017). An overview of multi-task learning in deep neural networks. arXiv:1706.05098.
60. Schroff, F., Kalenichenko, D., Philbin, J. (2015). Facenet: a unified embedding for face recognition and clustering. In *Proceedings of the IEEE conference on computer vision and pattern recognition* (pp. 815–823).
61. Sermanet, P., Eigen, D., Zhang, X., Mathieu, M., Fergus, R., LeCun, Y. (2013). Overfeat: integrated recognition, localization and detection using convolutional networks. arXiv:1312.6229.
62. Shi, W., Cao, J., Zhang, Q., Li, Y., Xu, L. (2016). Edge computing: vision and challenges. *IEEE Internet of Things Journal*, *3*(5), 637–646.
63. Simon, M., & Rodner, E. (2015). Neural activation constellations: unsupervised part model discovery with convolutional networks. In *Proceedings of the IEEE international conference on computer vision* (pp. 1143–1151).
64. Sochor, J., Herout, A., Havel, J. (2016). Boxcars: 3D boxes as cnn input for improved fine-grained vehicle recognition. In *Proceedings of the IEEE conference on computer vision and pattern recognition* (pp. 3006–3015).
65. Sochor, J., Spanhel, J., Herout, A. (2018). Boxcars: improving fine-grained recognition of vehicles using 3D bounding boxes in traffic surveillance. *IEEE Transactions on Intelligent Transportation Systems*, *20*(1), 97–108.
66. Specht, D.F. (1988). Probabilistic neural networks for classification, mapping, or associative memory. In *IEEE international conference on neural networks*, (Vol. 1 pp. 525–532).
67. Stark, M., Krause, J., Pepik, B., Meger, D., Little, J.J., Schiele, B., Koller, D. (2011). Fine-grained categorization for 3D scene understanding. *International Journal of Robotics Research*, *30*(13), 1543–1552.
68. Sun, W., Zhang, X., Shi, S., He, J., Jin, Y. (2017). Vehicle type recognition combining global and local features via two-stage classification. *Mathematical Problems in Engineering, 2017*.
69. Szegedy, C., Liu, W., Jia, Y., Sermanet, P., Reed, S., Anguelov, D., Erhan, D., Vanhoucke, V., Rabinovich, A. (2015). Going deeper with convolutions. In *Proceedings of the IEEE conference on computer vision and pattern recognition* (pp. 1–9).
70. Szegedy, C., Vanhoucke, V., Ioffe, S., Shlens, J., Wojna, Z. (2016). Rethinking the inception architecture for computer vision. In *Proceedings of the IEEE conference on computer vision and pattern recognition* (pp. 2818–2826).
71. Taek Lee, J., & Chung, Y. (2017). Deep learning-based vehicle classification using an ensemble of local expert and global networks. In *Proceedings of the IEEE conference on computer vision and pattern recognition workshops* (pp. 47–52).
72. Theagarajan, R., Pala, F., Bhanu, B. (2017). EDen: ensemble of deep networks for vehicle classification. In *Proceedings of the IEEE conference on computer vision and pattern recognition workshops* (pp. 33–40).
73. Wang, J., Yang, J., Yu, K., Lv, F., Huang, T., Gong, Y. (2010). Locality-constrained linear coding for image classification. In *2010 IEEE computer society conference on computer vision and pattern recognition* (pp. 3360–3367): Citeseer.
74. Wang, Y., & Yao, Q. (2019). Few-shot learning: a survey. arXiv:1904.05046.
75. Wieczorkowska, A., Kubera, E., Słowik, T., Skrzypiec, K. (2015). Spectral features for audio based vehicle identification. In *International workshop on new frontiers in mining complex patterns* (pp. 163–178): Springer.
76. Xiang, Y., Fu, Y., Huang, H. (2019). Global topology constraint network for fine-grained vehicle recognition. *IEEE Transactions on Intelligent Transportation Systems*.
77. Xiao, T., Xu, Y., Yang, K., Zhang, J., Peng, Y., Zhang, Z. (2015). The application of two-level attention models in deep convolutional neural network for fine-grained image classification. In *Proceedings of the IEEE conference on computer vision and pattern recognition* (pp. 842–850).
78. Xu, K., Ba, J., Kiros, R., Cho, K., Courville, A., Salakhudinov, R., Zemel, R., Bengio, Y. (2015). Show, attend and tell: neural image caption generation with visual attention. In *International conference on machine learning* (pp. 2048–2057).
79. Xu, Z., Yang, W., Meng, A., Lu, N., Huang, H., Ying, C., Huang, L. (2018). Towards end-to-end license plate detection and recognition: a large dataset and baseline. In *Proceedings of the European conference on computer vision (ECCV)* (pp. 255–271).
80. Yang, L., Luo, P., Change Loy, C., Tang, X. (2015). A large-scale car dataset for fine-grained categorization and verification. In *Proceedings of the IEEE conference on computer vision and pattern recognition* (pp. 3973–3981).
81. Zeng, R., Ge, Z., Denman, S., Sridharan, S., Fookes, C. (2019). Geometry-constrained car recognition using a 3D perspective network. arXiv:1903.07916.
82. Zheng, H., Fu, J., Mei, T., Luo, J. (2017). Learning multi-attention convolutional neural network for fine-grained image recognition. In *Proceedings of the IEEE international conference on computer vision* (pp. 5209–5217).
83. Zhong, Z., Zheng, L., Kang, G., Li, S., Yang, Y. (2017). Random erasing data augmentation. arXiv:1708.04896.

**Publisher's Note** Springer Nature remains neutral with regard to jurisdictional claims in published maps and institutional affiliations.

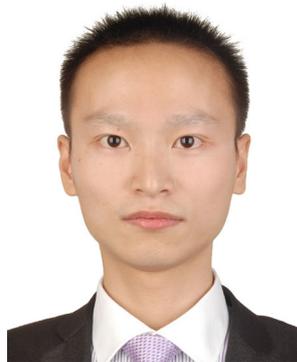

**Xingyang Ni** received his MSc degree in Information Technology from Tampere University of Technology in 2016. Currently, he is pursuing a PhD degree in Computer Vision at Tampere University. He has contributed to various projects on Deep Learning, in both academia and industry. His research interests include person re-identification, feature learning and image retrieval.

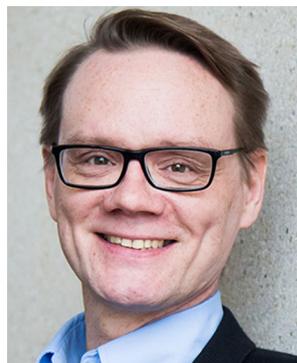

**Heikki Huttunen** received his PhD Degree in Signal Processing at Tampere University of Technology (TUT), Finland, in 1999. Currently he is an associate professor at the Unit of Computing Sciences at Tampere University, and involved in industrial development projects on automated image analysis and pattern recognition. He is an author of over 100 research articles. His research interests include Optical Character Recognition (OCR), Computer vision, Pattern recognition and Statistics.